\documentclass[runningheads]{llncs}
\usepackage[utf8]{inputenc} 
\usepackage[T1]{fontenc}    
\usepackage{hyperref}       
\usepackage{url}            
\usepackage{booktabs}       
\usepackage{amsfonts}       
\usepackage{nicefrac}       
\usepackage{microtype}      
\usepackage{lipsum}		
\usepackage{graphicx}
\usepackage{doi}
\usepackage{subcaption}
\begin{document}

\title{Generative Adversarial Networks for Astronomical Images Generation}
%
%
\author{Davide Coccomini\inst{1} \and
Nicola Messina\inst{2} \and
Claudio Gennaro\inst{2} \and Fabrizio Falchi\inst{2}}

\authorrunning{D. Coccomini et al.}
%
\institute{University of Pisa, Pisa, Italy
\email{d.coccomini@studenti.unipi.it}\\
\and
ISTI-CNR, 
via G. Moruzzi 1, 56124,
Pisa, Italy
\email{nicola.messina@isti.cnr.it, claudio.gennaro@isti.cnr.it, fabrizio.falchi@isti.cnr.it}}
\maketitle              
\begin{abstract}
Space exploration has always been a source of inspiration for humankind, and thanks to modern telescopes, it is now possible to observe celestial bodies far away from us. With a growing number of real and imaginary images of space available on the web and exploiting modern Deep Learning architectures such as Generative Adversarial Networks, it is now possible to generate new representations of space. In this research, using a Lightweight GAN, a dataset of images obtained from the web, and the Galaxy Zoo Dataset, we have generated thousands of new images of celestial bodies, galaxies, and finally, by combining them, a wide view of the universe.
The code for reproducing our results is publicly available at https://github.com/davide-coccomini/GAN-Universe, and the generated images can be explored at https://davide-coccomini.github.io/GAN-Universe/.

\end{abstract}


\section{Introduction}
Generative Adversarial Networks (GANs) \cite{goodfellow2014generative} have recently achieved astonishing results on many image generation tasks \cite{karras2020training}, producing figures that are difficult to distinguish from real ones in many different contexts.
At the same time, space exploration has enabled us to observe the universe with increasing clarity and detail, as in the case of the Hubble telescope, which was able to take an image of a portion of the universe from the combination of multiple individual photographs or the first-ever picture of a black hole obtained in 2019\footnote{https://www.nationalgeographic.com/science/article/first-picture-black-hole-revealed-m87-event-horizon-telescope-astrophysics}. Galloping technology is also giving rise to so-called space tourism with billionaires and ordinary people making space flights. All this is helping to fascinate new generations with the wonders of the universe and inspiring the companies to progress faster and faster in this area \cite{crawford2016longterm}. So we are now faced with an extremely high level of interest in space exploration, and curiosity about the wonders out there is higher than ever.

In this research, we aim to exploit Generative Adversarial Networks to generate new images of celestial bodies (planets, stars, galaxies, nebulae, etc.) based on actually captured photographs of the universe or  artistic productions available on the web. We will then also exploit the Galaxy Zoo dataset \cite{Willett_2013} containing hundreds of thousands of photographs of real galaxies to generate new ones and combine them in a wide view of the universe, similar to the Hubble image processing pipeline.

\section{Related Works}

\subsection{Generative Adversarial Networks}
Generative Adversarial Networks were first introduced by \cite{goodfellow2014generative}, and they immediately obtained a great success, being considered one of the biggest breakthroughs in the history of AI. GANs employ two distinct networks. The discriminator, the one that must be able to identify when an image is fake or not, and the generator, the network that actually generates new images in a sufficiently credible way to deceive its counterpart. With this technique, it was possible to generate new works of art \cite{elgammal2017can}, improve the resolution of existing photographs \cite{zhang2021ranksrgan} and even create highly credible deepfake images and videos \cite{Mirsky_2021}.
StyleGAN2 currently represents the state-of-the-art of these networks, with its adaptive discriminator augmentation (ADA) \cite{karras2020training}. The authors proposed a particular mechanism of data augmentation that significantly stabilizes training in limited data regimes, reducing the possibility of discriminator overfitting with the  consequent divergence of the training process. This is a big step forward because this new structure makes it possible to achieve excellent results even with limited data.
A further alternative for working in limited data situations was taken with the Lightweight GAN \cite{liu2021faster}, a further simplified version of this model capable of achieving good results but with a lighter and shorter training process. To do that, the architecture uses a skip-layer channel-wise excitation module and a self-supervised discriminator trained as a feature-encoder.  

\subsection{Previous applications of GANs in Astronomy}
GANs have been used in astronomy before, for example, to retrieve features in astrophysical images of galaxies with the so-called GalaxyGAN \cite{Schawinski_2017}. Very often, astrophysical images are disturbed by noise; deconvolution techniques are traditionally used to improve the quality of these observations, but these are rather limited. Using GANs, it is possible to recover lost features obtaining more faithful and reliable results.
In \cite{Smith_2019}, the authors exploited GANs, particularly Spatial-GANs \cite{sgan}, to generate views of space and thus of several celestial bodies in the same area, inspired by the data collected by the Hubble Space Telescope, obtaining quite realistic results. Using chained Generative Adversarial Networks \cite{Fussell_2019}, they were able to generate new galaxies with the same physical properties, highlighting how these architectures can be useful for data augmentation in this area as well.
GANs in this sector have also been useful for atmospheric retrievals on exoplanets to obtain a model, ExoGAN \cite{Zingales_2018}, capable of carrying out this task with a lower computational cost than traditional ones and able to recognize molecular features, atmospheric trace-gas abundances, and planetary parameters. 

\section{Methodology}
To train the network to generate celestial bodies, it is necessary to create a sufficiently large and heterogeneous dataset. To do this, we collected both real space images and artistic representations of the universe using the Flickr Scraper library\footnote{\url{https://github.com/ultralytics/flickr_scraper}}, together with some handwork for manually downloading images from the web. The collected dataset was then manually revised to discard inconsistent images (e.g., the trivial ones or those having too low resolution). As the images collected online were all of different sizes, they were then cropped in a centered squared way, ensuring that the image was not deformed during the resize phase. In the end of this process, we obtained a dataset of 283 coherent, good quality, and squared images.

\begin{figure}[t]
\begin{subfigure}[b]{0.5\textwidth}
\centering
\includegraphics[width=\linewidth]{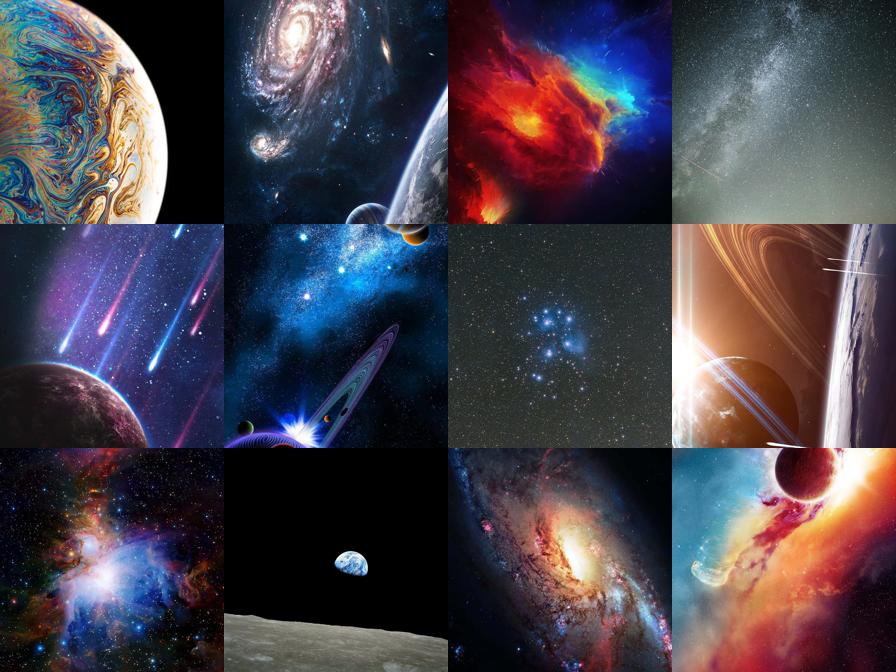}
\caption{Examples of collected dataset images.}
    \label{figure:dataset}
\end{subfigure}
\hspace{0.5cm}
\begin{subfigure}[b]{0.5\textwidth}
\centering
\includegraphics[width=\linewidth]{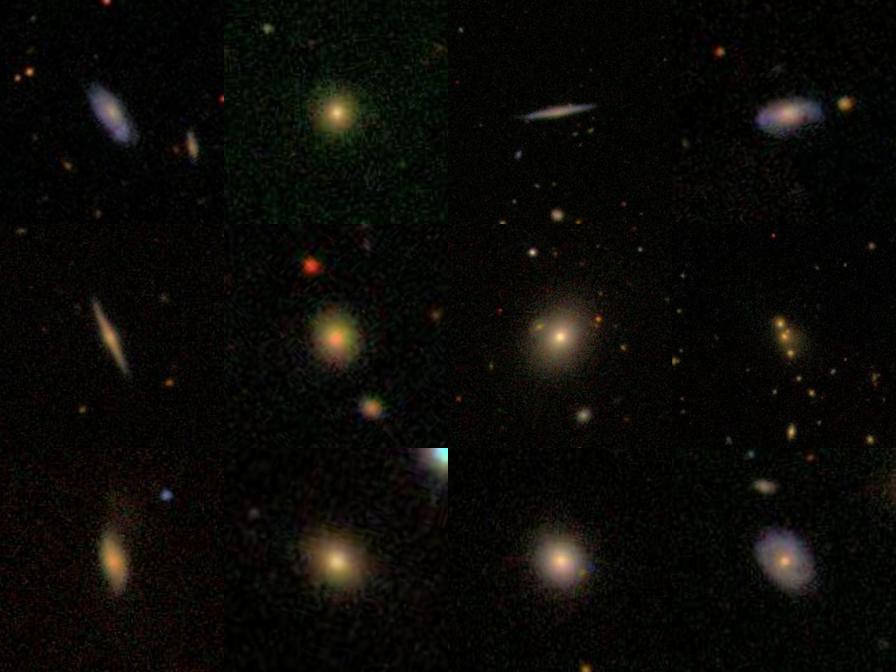}
\caption{Examples of Galaxy Zoo Dataset images.}
\label{figure:galaxyzoo}
\end{subfigure}
\caption{Examples of images in the dataset.}
\end{figure}

We also exploited the Galaxy Zoo Dataset \cite{Willett_2013}, a large collection with hundreds of thousands of space images collected by telescopes, to carry out further tests and obtain real galaxy images to be merged into a single wide view.

The Generative Adversarial Network chosen to carry out the experiments is a Lightweight GAN \cite{liu2021faster}, a version very similar to the state-of-the-art StyleGAN2 but lighter and easier to train. 
In fact, it has been demonstrated that this network is able to converge in a few hours, on a single GPU, with a few hundred training samples and achieving remarkable quality results. For these reasons, it is the more suitable architecture in our context.

\section{Experiments}
We trained two different instances of the Lightweight GAN using first our collected data and then the one from the Galaxy Zoo Dataset. In both cases, the only data augmentation technique applied was to vary the color of the input images with a probability of 25\%.

\subsection{Training}
Firstly, a Lightweight GAN was trained on the 283 images collected from the web. Subsequently, another instance of the network was trained on a portion of the Galaxy Zoo Dataset consisting of 61636 images. In both cases, we used a batch size of 3 with an Adam optimizer and a learning rate of $2e^{-4}$ . The generated images are 128x128 pixels for the first dataset, to simplify the network's training process with such a low number of input images, and 256x256 for the second one, in which we have a larger number of available images.

\subsection{Results}
\begin{figure}[t]
\begin{subfigure}[b]{0.5\textwidth}
\centering
\includegraphics[width=\linewidth]{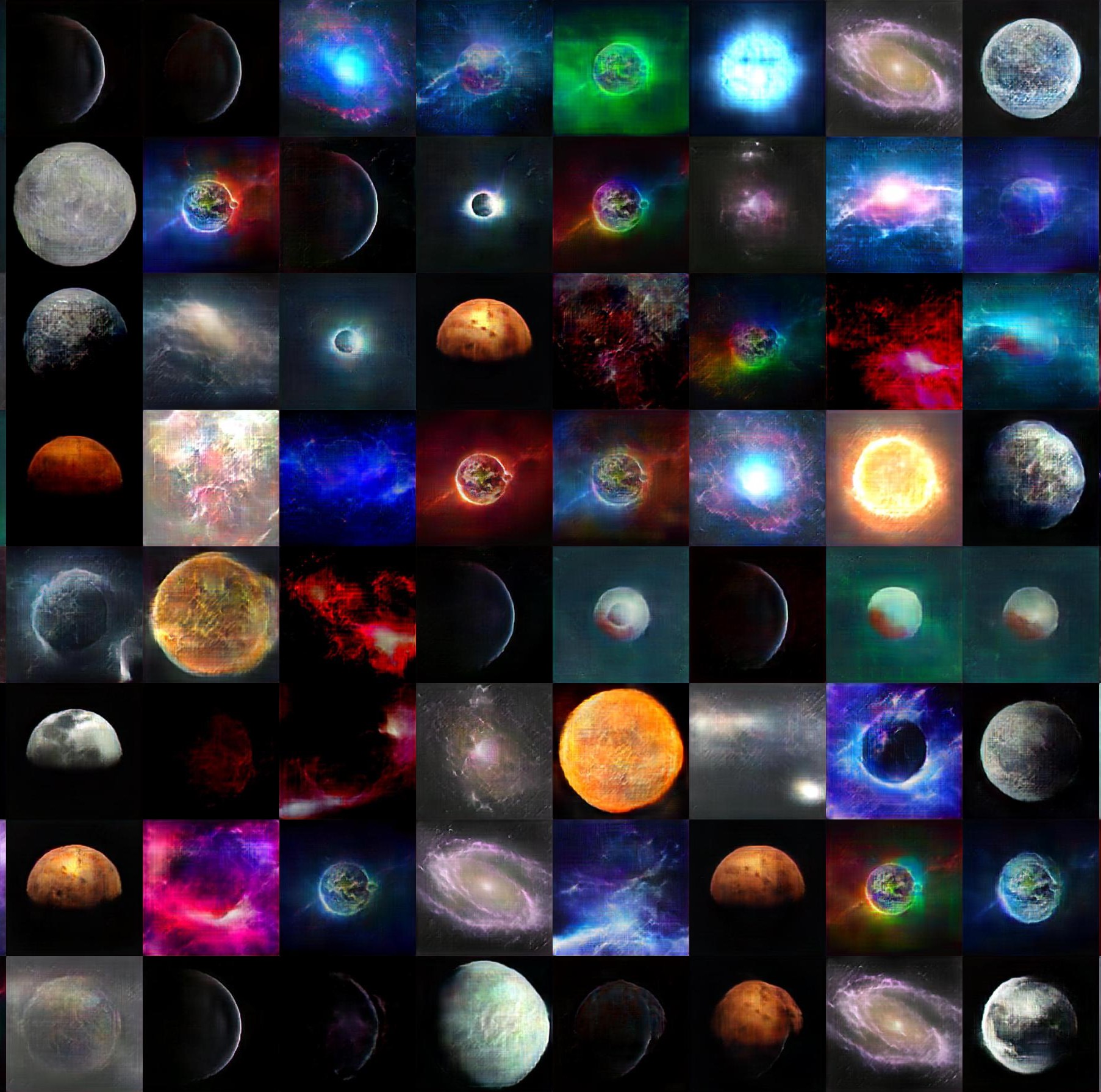}
\caption{Examples of generated images from Lightweight GAN trained on scraped images from web.}
    \label{figure:result1}
\end{subfigure}
\hspace{0.5cm}
\begin{subfigure}[b]{0.5\textwidth}
\centering
\includegraphics[width=\linewidth]{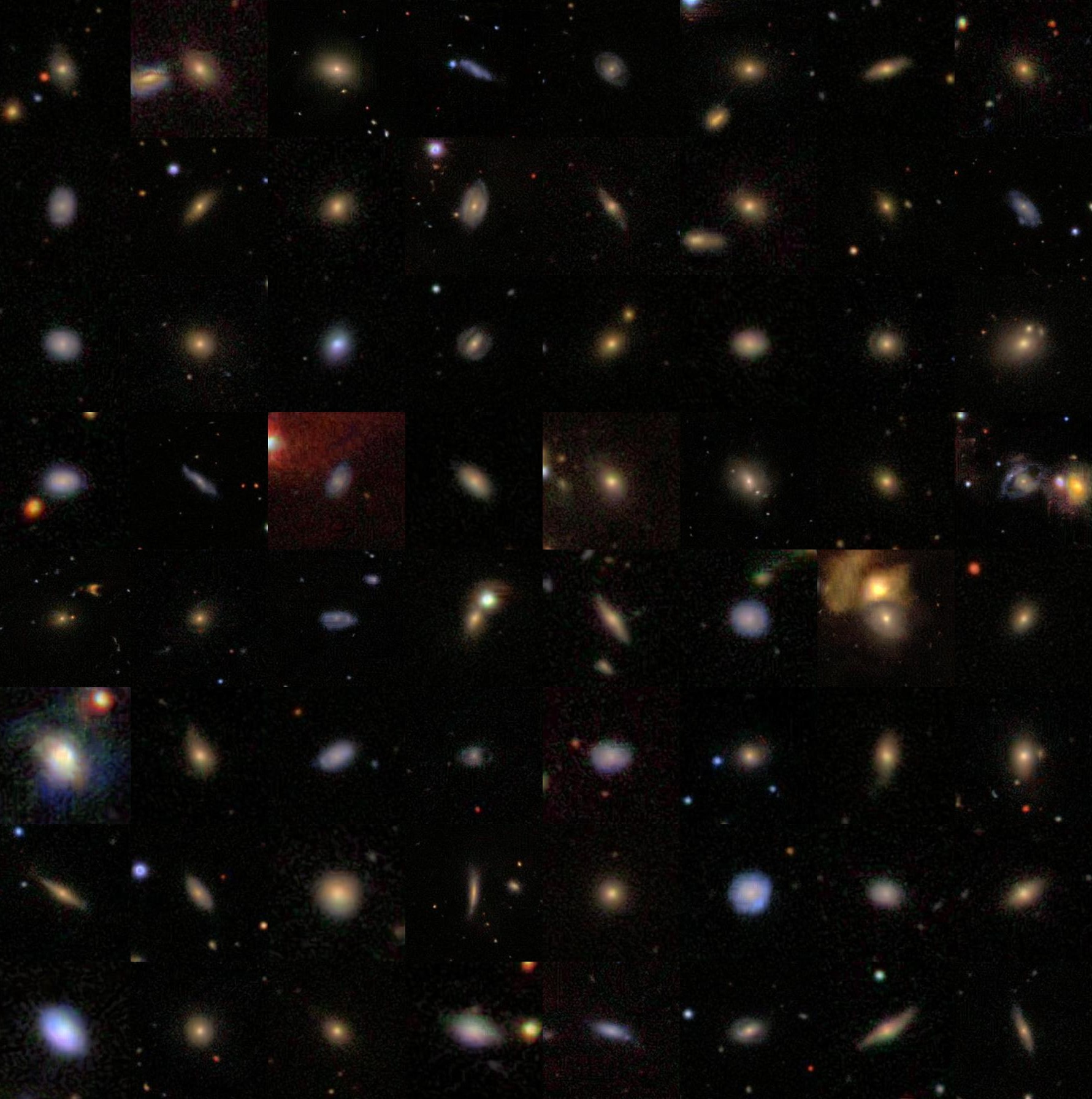}
\caption{Examples of generated images from Lightweight GAN trained on Galaxy Zoo Dataset.}
\label{figure:result2}
\end{subfigure}
\caption{Examples of images generated by the GAN.}
\end{figure}

The lightweight GAN trained on the celestial bodies dataset obtained from the web produced on a single Tesla T4 GPU, after three days of training, some excellent and very credible images of planets, nebulae, and galaxies, with style very similar to that provided as input as shown in Figure \ref{figure:result1}.

After two days of training in the same environment, we obtained very credible images  on the Galaxy Zoo Dataset. In this case, thanks to the simplicity of the input images and their abundance, the trained network could produce results hardly distinguishable from the original ones, as shown in Figure \ref{figure:result2}.

\begin{figure}[t]
    \centering
    \includegraphics[width=1\textwidth]{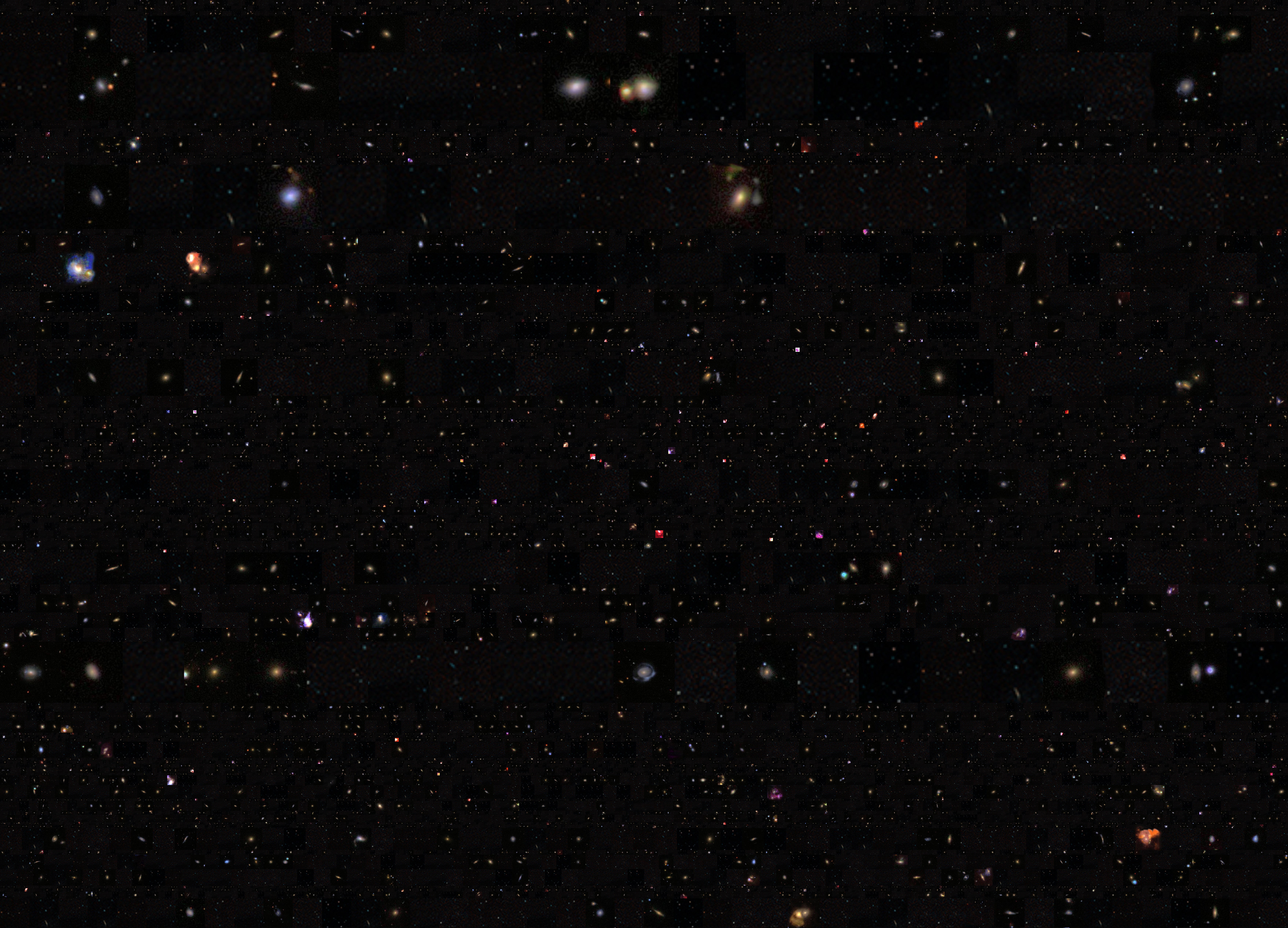}
    \caption{Wide view of universe generated by Lightweight GAN.}
    \label{figure:result3}
\end{figure}

Finally, inspired by the work done by astronomers who in 2019 used 7500 individual exposures obtained by the Hubble telescope to create a mosaic of a wide view of the universe\footnote{https://www.nasa.gov/feature/goddard/2019/hubble-astronomers-assemble-wide-view-of-the-evolving-universe}, we exploited the Lightweight GAN trained on the Galaxy Zoo dataset to generate a wide view of the universe. We first obtained 3000 images of galaxies from the network, and a further 10, extremely simple and almost totally black, were selected as blank space. These images were rotated, resized, and randomly combined to obtain a mosaic of 25000 images, thus generating the desired wide view, which can be seen in Figure \ref{figure:result3}.

Although the obtained galaxies may look very similar to the original ones to the naked eye, we decided to conduct an additional experiment using a pre-trained detector trained on Galaxy Zoo to get a more quantitative evaluation. The network chosen is the one trained by \cite{gonzalez2018galaxy} and based on YOLOv2. By looking at the detection results from this network, we can get a more objective idea of the results obtained.

As shown in Figure \ref{figure:galaxy_prediction_full}, using as input the wide-view generated by the combination of galaxies obtained from our GAN, the detector is able to identify many of the larger galaxies in the scene.
 Repeating the test on zoomed portions of the image, the detector continues to detect a good number of galaxies perfectly in line with the tests conducted in the paper.

To obtain a quantitative measure of the quality of the images generated by the GAN, we used  the Fréchet Inception Distance (FID) \cite{heusel2018gans}. This score is a widely used metric to assess the quality of a GAN's output. It gives an assessment based on both the variety of images generated and their credibility taking into account the classes distribution from the training set. The network chosen to calculate this metric is an XceptionNet developed for Kaggle Galaxy Zoo competition\footnote{https://www.kaggle.com/c/galaxy-zoo-the-galaxy-challenge} and publicly available\footnote{https://www.kaggle.com/hironobukawaguchi/galaxy-zoo-xception/notebook} trained on 61577 images from GalaxyZoo training set to classify into three distinct classes, spiral, eliptical and irregular. The training images were used together with 3000 images generated by GAN to obtain the Fréchet Inception Distance value. The FID obtained is 0.007, very close to 0, indicating a good fidelity of the generated images with respect to the original ones. The distribution is also very similar, as can be seen in the Table \ref{tab:distibution}:

\begin{figure}[t]
    \centering
    \includegraphics[width=1\textwidth]{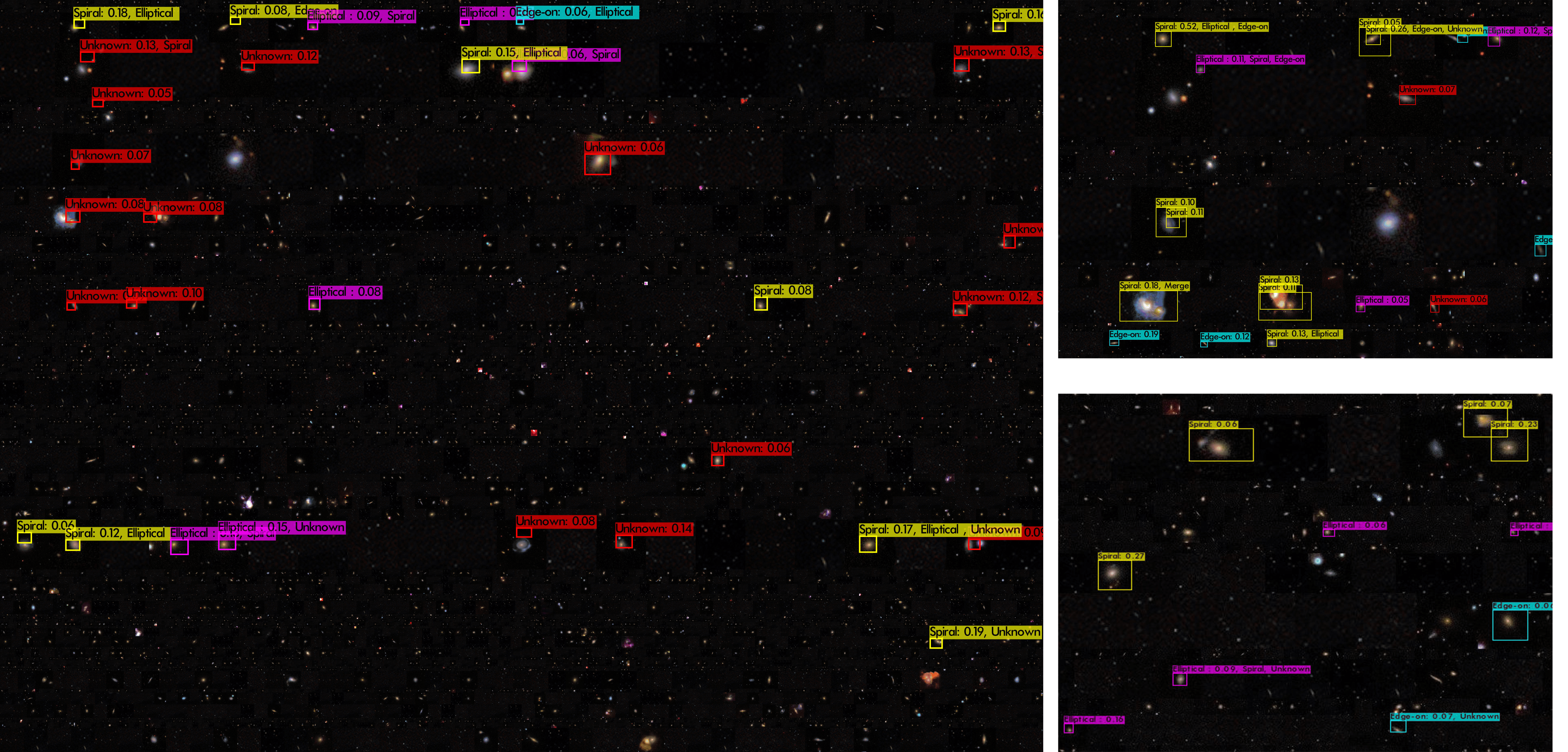}
    \caption{Galaxies detection on the entire image.}
    \label{figure:galaxy_prediction_full}
\end{figure}
\begin{table}[t]
\caption{Classes distribution on training and generated sets}
\label{tab:distibution}
\centering
\resizebox{0.7\textwidth}{!}{
\begin{tabular}{lccc} 
    \toprule
    \textbf{Set}    & Spiral & Eliptical & Irregular \\
    \midrule
    Training       & 25775 (41.8\%)  &  35738 (58.0\%) & 65 (0.1\%)\\
    Generated      & 920 (30.6\%) & 2064 (68.8\%) & 16 (0.5\%) \\
   
    \bottomrule
\end{tabular}
}
\end{table}
However, an FID value so close to zero and therefore so good could also result from a GAN memorizing the examples shown to it during training, a danger highlighted in \cite{lucic2018gans}. In order to rule out this possibility, we carried out a similarity search between 40 images randomly extracted from those generated by GAN and all the images in the training set. 

\begin{figure}[t]
    \centering
    \includegraphics[width=0.85\textwidth]{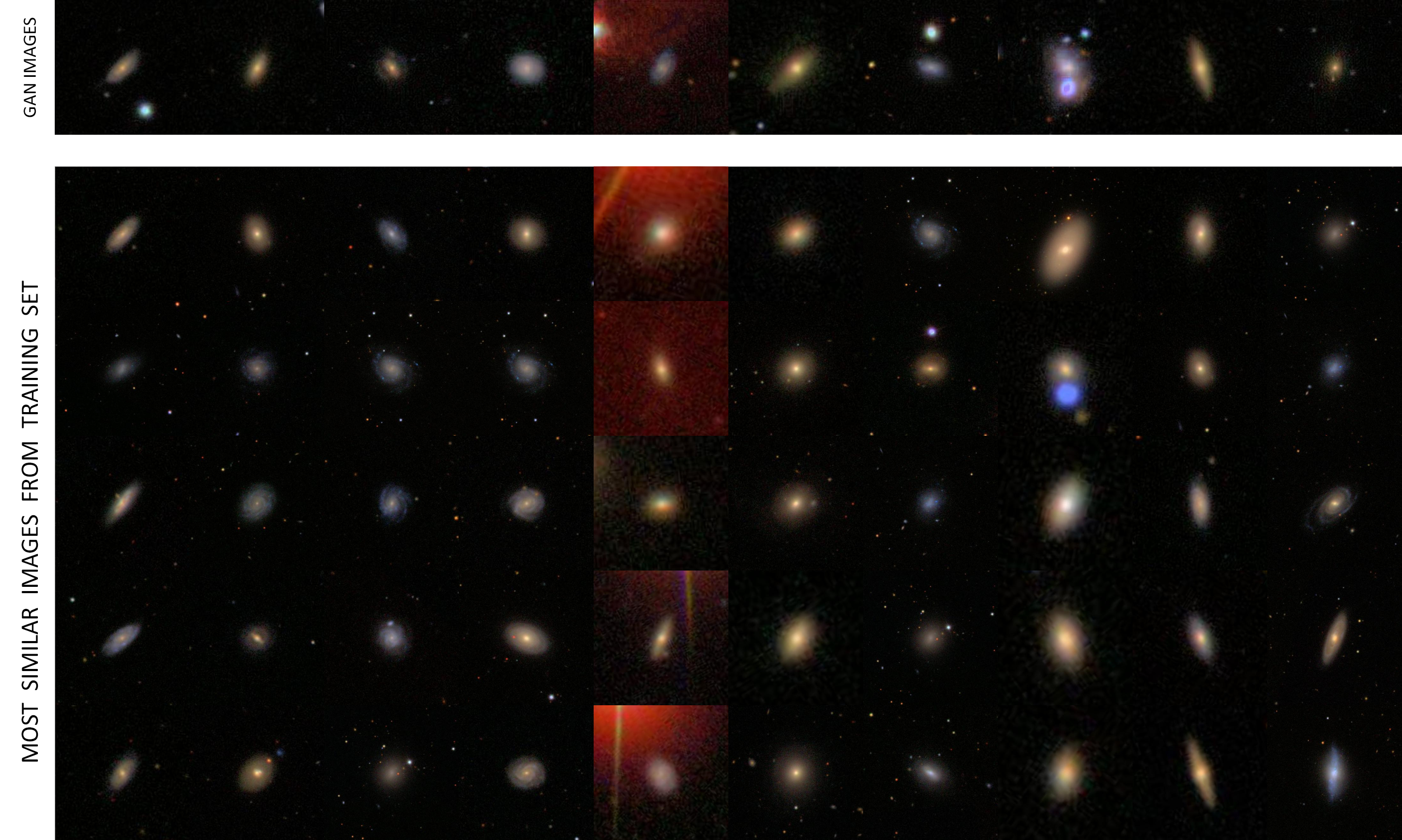}
    \caption{The figure shows 10 images generated by the GAN training (first row), and for each of them, on the same column, the five most similar images identified within the training set. }
    \label{figure:compare}
\end{figure}

The comparison of each image pair (fake/real) was conducted by calculating the Structural Similarity Index Measure (SSIM), a commonly used measure of similarity between two images.  As shown in Figure \ref{figure:compare}, in none of the cases considered was an image from the training set found to be perfectly identical to the one generated, but it is often very clear that the image is completely new. 
In the end, the conducted experiments show good results in both qualitative and quantitative terms.

\section{Conclusions}
In this research we demonstrated the ability of a Lightweight GAN to obtain extremely credible results even with a low amount of data and in an almost unexplored context. We also generated a multitude of credible images of a wide variety of individual celestial bodies, galaxies, and finally, a spectacular wide-view of a portion of the universe obtained from the combination of the created images. The quality of the results has been validated in multiple ways: (a) exploiting purely aesthetic evaluations and detection techniques using pre-trained networks and (b) measuring objective metrics and parameters widely used within the  Generative Adversarial Networks framework.
The resulting network can be used to generate images useful in the world of art and graphics and as a data augmentation tool for the classification of planets or galaxies, once again highlighting the usefulness and effectiveness of Generative Adversarial Networks.


\bibliographystyle{splncs04}
\bibliography{references}
\end{document}